\pdfoutput=1

\documentclass[11pt]{article}

\usepackage[final]{acl}

\usepackage{times}
\usepackage{latexsym}

\usepackage[T1]{fontenc}

\usepackage[utf8]{inputenc}

\usepackage{microtype}

\usepackage{inconsolata}

\usepackage{graphicx}

\usepackage{multirow}
\usepackage{colortbl}
\usepackage{bbm}

\usepackage{xcolor}
\usepackage{nicematrix}

\title{Is Large Language Model Performance on Reasoning Tasks \\ Impacted by Different Ways Questions Are Asked?}

\author{Seok Hwan Song, Mohna Chakraborty, Qi Li, Wallapak Tavanapong \\ Department of Computer Science, Iowa State University, Ames, Iowa, USA \\ \texttt{\{song92, mohnac, qli, tavanapo\}@iastate.edu}}

\begin{document}

\maketitle
\begin{abstract}
Large Language Models (LLMs) have been evaluated using diverse question types, e.g., multiple-choice, true/false, and short/long answers. This study answers an unexplored question about {\em the impact of different question types on LLM accuracy on reasoning tasks. }We investigate the performance of five LLMs on three different types of questions using quantitative and deductive reasoning tasks. The performance metrics include accuracy in the reasoning steps and choosing the final answer. \textbf{Key Findings:} (1) Significant differences exist in LLM performance across different question types. (2) Reasoning accuracy does not necessarily correlate with the final selection accuracy. (3) The number of options and the choice of words, influence LLM performance. 

\end{abstract}

\section{Introduction}

Performance of Large Language Models (LLMs) on reasoning tasks has been extensively studied using diverse benchmarks \citep{weston2016babi, gsm, huang-chang-2023-towards}. Several benchmarks use short answer questions (SAQs) where the LLMs generate a response, often accompanied by an explanation. Multiple-choice questions (MCQs) are widely used because they allow for simple assessment \citep{order}. LLMs select the correct option among the given options. For True or False questions (TFQs), LLMs judge if the given statement/question is true or false. Factors that may influence LLM performance include the number of options, the sensitivity of word choices in MCQs, and whether ``True'' or ``False'' is the correct answer in TFQs.

Assessing LLM performance can be challenging due to the diverse question types. SAQs have the advantage of analyzing reasoning steps with clear answers. For MCQs, LLMs need to understand the options correctly and choose the right option. LLMs may overlook the reasoning steps. TFQs require LLMs to understand true or false accurately. 

Motivated by these challenges, we evaluate LLM performance on reasoning questions across different question types on two reasoning tasks, {\it quantitative reasoning task and a deductive reasoning task.} Quantitative reasoning datasets were evaluated to benchmark multiple existing LLMs \citep{gpt4, llama3, gemma}, and it contains arithmetic calculation. We also chose deductive reasoning tasks to evaluate LLMs' logical reasoning without arithmetic calculations.

Our study focuses on assessing the suitability of these question types for benchmarking purposes and gaining insights into the behavior of LLMs. We propose {\it final selection accuracy} and {\it reasoning accuracy} for performance assessment. Final selection accuracy evaluates only the final answer in the output, such as the selected option in MCQs or TFQs. Reasoning accuracy assesses the correctness of the reasoning steps leading to the final answer. Checking reasoning accuracy is time-consuming since it cannot be automated.
Consequently, we introduce the following research questions:

\begin{enumerate}
\item (RQ1) Do question types (SAQs, MCQs, and TFQs) impact the final selection accuracy of LLM? 
\item (RQ2) Do question types impact the reasoning accuracy of LLM? 
\item (RQ3) What are the patterns of incorrect outputs by LLMs on different question types?
\item (RQ4) What factors of MCQs impact LLM performance?
\item (RQ5) What factors of TFQs influence LLM performance?
\end{enumerate}


{\bf Contribution \#1:} The aforementioned research questions and the experimental design to answer the questions. We carefully designed the options for MCQs and TFQs. To our best knowledge, RQ1, RQ2, RQ3, and RQ5 have not been investigated. RQ4 was partially investigated in \citep{selection} but with the following differences: (1) different numbers of options, (2) the wrong options in our MCQs covering a wide range of mistakes, and (3) the inclusion of the ``Something else" option that enlarges the solution space in reasoning in our MCQs.


{\bf Contribution \#2:} New key findings from evaluations of five LLM models, two closed-source LLMs, and three open-source LLMs on quantitative reasoning problems selected from the GSM8K dataset~\citep{gsm} and deductive reasoning problems selected from the bAbI dataset~\citep{weston2016babi}. We will share our code and benchmark publicly\footnote{Code and benchmark: \url{https://github.com/NRT-D4/LLM-bias-questiontype}}. Some key findings are as follows. (1) There are more statistically significant differences in the final accuracy across different question types compared to the reasoning accuracy. (2) Reasoning accuracy does not necessarily correlate with final selection accuracy. (3) LLMs perform better with ``True'' as correct answers than ``False'' in TFQs. For MCQs, the LLM performances vary significantly based on the number of questions. The insight is potentially useful to guide future benchmark development and improve LLM performance.

\section{Related Work \label{related_work}}

\begin{figure*}[h!tb] {
\centering
\includegraphics[width=1\textwidth]{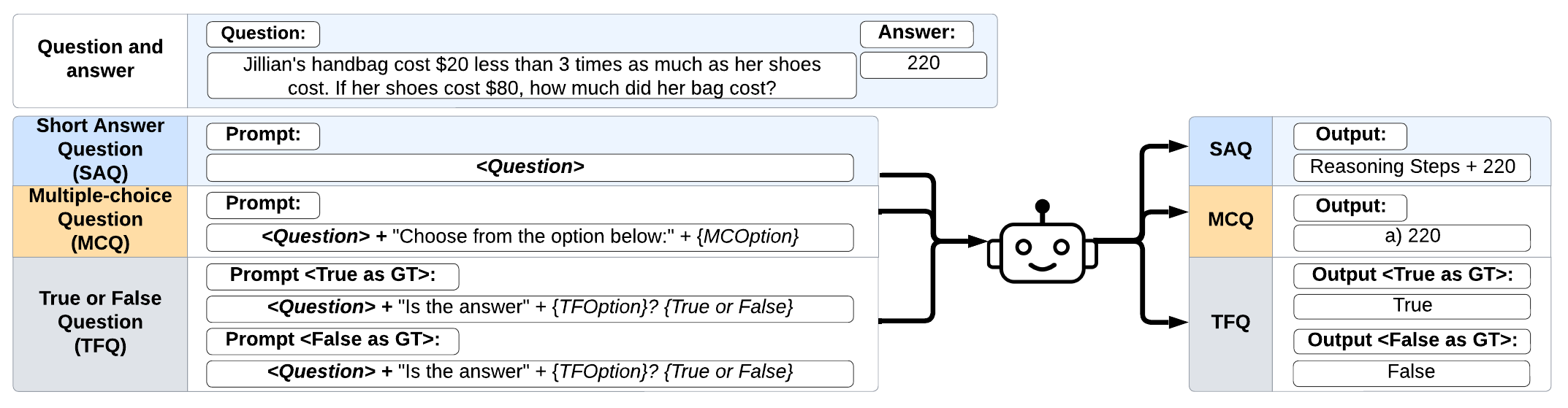} 
\caption{Examples of different types of questions generated from the original problem. Variables in italics are replaced by actual values; ``+'' indicates concatenation. {\it MCOption} represents multiple choices options; {\it TFOption} is either a correct or incorrect answer for TFQs.}
\label{qtypes}}
\end{figure*}

Diverse benchmarks \citep{gsm, svamp} were used to evaluate LLMs \citep{gpt4, llama3, gemma} and prompting methods \citep{scott}. Several datasets \citep{hendryckstest2021, ARC2021, sakaguchi-winnogrande-2021} include MCQs \citep{selection}. Moreover, factors influencing performance on MCQs have been studied, for instance, selection biases by order of options \citep{selection, order, first}, and types of token options \citep{selection}. \citet{preknow} found that LLMs can solve MCQs without knowing the question after some few-shot examples. \citet{order} claims that the LLM performance of MCQs is better than those of long-form generation questions on the CARE-MI dataset. They do not find a significant difference in the different number of options for MCQs. In their study, the number of options varies from 2 to 4 without ``Something else'' as a possible answer. Our MCQs contain 5 options and 11 options with ``Something else'' as the final answer.

\section{Benchmark Construction}

Our goals for designing the benchmark include (1) ensuring that multiple-choice and True-or-False questions are as difficult as short-answer questions in terms of reasoning; (2) evaluating LLM accuracy in both choosing the final answers and its reasoning steps; and (3) answering the three research questions while avoiding the bias issues mentioned in the related work.

\subsection{Reasoning Tasks and Source Datasets}

\begin{itemize}
\item \textbf{Quantitative reasoning with GSM8K \citep{gsm}:} GSM8K is one of the standard datasets used for evaluating LLM quantitative reasoning performance. The dataset has grade school-level math word problems in short-answer question types. For each problem, the answer and steps in deriving the answer are included.

\item \textbf{Deductive reasoning with bAbI-Factoid QA with three supporting facts:} This dataset is a subset of the bAbI dataset \citep{weston2016babi}. Each problem has three supporting facts in the short-answer question type. 

\end{itemize}    

We randomly selected 300 reasoning problems from each dataset, considering the API budget for the closed-source LLMs, and the time-consuming manual annotation to evaluate LLM reasoning performance. To add reasoning complexity to deductive reasoning tasks, for each selected problem, we added two additional facts irrelevant to the question of the problem. These facts were randomly selected from the other questions in the bAbI dataset. We denoted the dataset with the 300 selected GSM8K problems as GSM8K300 and the modified bAbI dataset as bAbI300. 
See examples in A.3.


\subsection{Prompt Designs to Format Question Types}

 Given each reasoning problem, different prompt structures are used to format the problem into corresponding SAQ, MCQ, and TFQ question types. Figure \ref{qtypes} shows our example prompts. 

\subsubsection{Short Answer Questions (SAQs)}

The prompt structure is simple. It only has the question without any other instructions to LLMs. They return their reasoning steps and the final answer.

\subsubsection{Multiple Choice Questions (MCQs) \label{mcq-design}}

This type of question contains a question, a brief instruction, and a list of options. Each option contains a token (e.g., ``a)'', ``b)'') followed by a value. We investigate several design factors but fix the token format since previous work has shown that the token formats (i.e., a/b/c/d, 1/2/3/4, and (A)/(B)/(C)/(D)) do not result in statistically significant differences in LLM performance~\citep{selection}. 

{\bf Number of options:} Previous work found that LLM performances on 2, 3, and 4 option MCQs are not statistically significant~\citep{order}. We created 5-option and 11-option MCQs. The 11-option MCQs are rare but included in the Numersense dataset~\citep{numersenselin-etal-2020-birds}. We aim to test whether the significant difference in the number of options leads to observed notable performance variations. If LLMs perform actual calculations through reasoning steps, the number of options should not matter. However, if LLMs guess or test each option step-by-step, the chance of guessing correctly with eleven options is significantly less. 

{\bf ``Something else'' or (SEO) as the last option value:} \label{something} Our rationale is to ensure that LLMs can select an option for their reasoning on our multiple-choice questions when SEO is included as an option, just like in short-answer questions.
 
{\textbf{Positions of correct answers:} A multiple-choice question has only one correct option among all positions, including the final option ``Something else.''} LLMs were found to have {\it selection bias} defined as the bias to select specific options as
answers \citep{selection}. We investigate three approaches for placing the correct answers.
\begin{enumerate}
\item \textbf{Uniform across all positions (U):} Across all the problems, each option, including SEO, has an equal chance of being the correct answer.  For instance, for the dataset with 300 reasoning problems, with five-option multiple-choice questions, 60 problems have the first option as the correct answer; 60 other problems have the correct answers in the second option, and so on.
\item \textbf{Uniform across all non-SEO positions (U-SEO):} Each option, except SEO, has an equal chance of being the correct answer. For the same example, the first option is the correct answer for 75 problems. The second option is the correct answer for 75 other problems, and so on. SEO positioned at the end has an incorrect answer.
\item \textbf{Only SEO (oSEO):} The correct answer is SEO for all the reasoning problems.
\end{enumerate}

{\bf Option values for wrong answers:} For the quantitative reasoning task, we designed three types of wrong options for each question. In Type 1, the wrong answer is 5\% different from the ground truth, whereas in Type 2, the wrong option is 20\% different. For Type 3, we used the numeric values from the question to fill in a generated equation with random operators and check that the result of the equation is not the same as the ground truth before using the result as the wrong option. When there were too few numeric values in a reasoning problem to generate different wrong answers, we manually created wrong answers.
A 5-option MCQ has three wrong options, one for each type, whereas an 11-option MCQ has three wrong options for each type. For the deductive reasoning task, the wrong option values were selected from a predefined list. For a reasoning problem that SEO is the correct answer, another wrong option value is needed. For quantitative reasoning, we randomly generated the wrong option value that has the same number of digits as the ground truth. For deductive reasoning, the wrong value was randomly selected from the predefined list.

\subsection{True or False Questions (TFQs) \label{tfq-design}}

This question type has either a correct or incorrect answer in the question. The output is either true or false. 
We designed two formats of TFQ questions, as shown below, to test whether LLMs are sensitive to how the questions are asked. 
We replaced \{answer\} and \{instruction\} variables with values for each reasoning problem when prompting LLMs.

\begin{itemize}
\item \textbf{Question format:}\\ ``Is the answer \{answer\}? + \{instruction\}''
\item \textbf{Statement format:}\\ ``The answer of the question is \{answer\}. + \{instruction\}''
\end{itemize}

The \{answer\} variable was substituted with the correct or incorrect answer. For GSM8K300, the incorrect answer was randomly picked from numbers with the same number of digits of the correct answer. For bAbI300, the incorrect answer was randomly selected from the list of options without the correct answer. We considered two different values for the \{instruction\} variable: (1)  
``True or False'' and (2) ``Solve the question first and choose True or False.'' We found that LLMs tend to select the true or false before solving the question. Therefore, we added the second instruction. We also studied the impact of using ``Yes or No'' instead of ``True or False'' on LLM performance.

\subsection{Evaluation Metrics}

We evaluate LLM performance using two metrics: final selection accuracy and reasoning accuracy.

\noindent \textbf{Final selection (FS) accuracy} is the ratio of the number of reasoning problems an LLM gives the correct answers to the total number of reasoning problems. The correct answer for each SAQ problem is the ground truth. For MCQs, the correct answer includes the correct token with the exception that if the correct answer is ``Something else,'' the token is not required. For TFQs, the final selection answer should simply be either ``True'' or ``False.'' Calculation of FS accuracy was automated using string matching between the outputs and the ground truths.

\noindent \textbf{Reasoning accuracy} is the ratio of the number of reasoning problems an LLM gives correct reasoning to the total number of reasoning problems. We used a strict criterion that the LLM needs to output all the reasoning steps correctly to be considered as giving the correct reasoning. This process cannot be automated. {\bf Manual Evaluation:} The first author trained three students. Each person evaluated distinct subsets of the complete LLM outputs group by the LLM model to keep the evaluations within the same model consistent. All evaluators communicated frequently to maintain evaluation consistency. The main author randomly verified the evaluation results. For SAQs, the reasoning accuracy is the same as the FS accuracy.

\begin{table*}[!htbp]
\centering
\small
\renewcommand{\arraystretch}{1.5} 
\setlength{\tabcolsep}{3pt} 
\begin{NiceTabular}{c|c|cccccc||cccccc}
\multirow{3}{*}{\rotatebox{90}{Accuracy}}      & \multirow{3}{*}{Models} & \multicolumn{6}{c||}{GSM8K300}                                                                                                                                              & \multicolumn{6}{c}{bAbI300}                                                                                                                                               \\ \cline{3-14} 
                                 &                         & \multicolumn{1}{c|}{\multirow{2}{*}{SAQ}} & \multicolumn{3}{c|}{MCQ (5 Options)}       & \multicolumn{2}{c||}{TFQ (Question)} & \multicolumn{1}{c|}{\multirow{2}{*}{SAQ}} & \multicolumn{3}{c|}{MCQ (5 Options)}       & \multicolumn{2}{c}{TFQ (Question)}                                                                   \\ \cline{4-8} \cline{10-14} 
                                 &                         & \multicolumn{1}{c|}{}                     & U & U-SEO & \multicolumn{1}{c|}{oSEO} & True & False & \multicolumn{1}{c|}{}                     & U & U-SEO & \multicolumn{1}{c|}{oSEO} & True & False \\\hline
\multirow{5}{*}{\rotatebox{90}{Final Selection (FS)}}  & GPT-4o                  & \multicolumn{1}{c|}{92.00}                                     & 89.45  & 91.33   & \multicolumn{1}{c|}{\cellcolor{pink}97.67}  & 93.00  & \cellcolor{pink}97.33*  & \multicolumn{1}{c|}{82.33}                                     & 79.00  & 80.00   & \multicolumn{1}{c|}{83.67}  & \cellcolor{cyan}67.67  & \cellcolor{pink}97.00$^\dagger$ \\\cline{2-14}
                                 & GPT-3.5                 & \multicolumn{1}{c|}{79.60}                                     & \cellcolor{cyan}69.67  & 83.67   & \multicolumn{1}{c|}{\cellcolor{cyan}16.67}  & 75.67  & \cellcolor{pink}96.67*  & \multicolumn{1}{c|}{52.67}                                     & \cellcolor{cyan}30.00  & \cellcolor{cyan}38.33   & \multicolumn{1}{c|}{\cellcolor{cyan}10.67}  & \cellcolor{cyan}16.33  & \cellcolor{pink}92.67$^\dagger$  \\\cline{2-14}
                                 & Gemma                   & \multicolumn{1}{c|}{58.67}                                     & 54.33  & \cellcolor{pink}67.33   & \multicolumn{1}{c|}{\cellcolor{cyan}2.00}  & \cellcolor{pink}98.33  & \cellcolor{cyan}10.00  & \multicolumn{1}{c|}{36.33}                                     & 39.33  & \cellcolor{pink}49.67   & \multicolumn{1}{c|}{\cellcolor{cyan}0.00}  & \cellcolor{pink}99.33  & \cellcolor{cyan}19.00  \\\cline{2-14}
                                 & Llama-8B                & \multicolumn{1}{c|}{75.33}                                     & \cellcolor{cyan}66.67  & 77.33   & \multicolumn{1}{c|}{\cellcolor{cyan}19.00}  & 73.67  & \cellcolor{pink}86.33*  & \multicolumn{1}{c|}{53.67}                                     & 59.00  & \cellcolor{pink}69.67   & \multicolumn{1}{c|}{\cellcolor{cyan}21.00}  & \cellcolor{pink}77.33  & \cellcolor{pink}88.00$^\dagger$  \\\cline{2-14}
                                 & Llama-1B                & \multicolumn{1}{c|}{48.00}                                     & \cellcolor{cyan}11.67  & \cellcolor{cyan}10.67   & \multicolumn{1}{c|}{\cellcolor{cyan}5.33}  & 47.67  & \cellcolor{cyan}37.00*  & \multicolumn{1}{c|}{37.67}                                        & \cellcolor{cyan}20.67  & \cellcolor{cyan}14.33   & \multicolumn{1}{c|}{\cellcolor{cyan}23.67}  & 36.33  & \cellcolor{cyan}19.33  
                                 \\\hline
                                 \hline
\multirow{5}{*}{\rotatebox{90}{Reasoning (R)}}       & GPT-4o                  & \multicolumn{1}{c|}{92.00}                                     & 93.67  & 93.67   & \multicolumn{1}{c|}{92.67}  & \cellcolor{pink}95.00  & \cellcolor{pink}94.00  & \multicolumn{1}{c|}{82.33}                                     & \cellcolor{cyan}74.33  & 81.67  & \multicolumn{1}{c|}{\cellcolor{cyan}41.81} & \cellcolor{cyan}59.33     & \cellcolor{cyan}68.33     \\\cline{2-14}
                                 & GPT-3.5                 & \multicolumn{1}{c|}{79.60}                                    & 81.00  & 80.67   & \multicolumn{1}{c|}{80.00}  & 78.00  & 79.33  & \multicolumn{1}{c|}{52.67}                                     & 53.33  & \cellcolor{cyan}60.67   & \multicolumn{1}{c|}{\cellcolor{cyan}37.67}  & 48.67  & 53.67  \\\cline{2-14}
                                 & Gemma                   & \multicolumn{1}{c|}{58.67}                                     & 54.85  & 60.67   & \multicolumn{1}{c|}{54.00}  & 57.00  & \cellcolor{cyan}45.67  & \multicolumn{1}{c|}{36.33}                                     & 38.00  & \cellcolor{pink}45.30   & \multicolumn{1}{c|}{\cellcolor{cyan}0.67}  & 30.67  & \cellcolor{cyan}2.67   \\\cline{2-14}
                                 & Llama-8B                & \multicolumn{1}{c|}{75.33}                                     & \cellcolor{pink}81.00  & \cellcolor{pink}81.00   & \multicolumn{1}{c|}{76.67}  & 76.67  & 77.33  & \multicolumn{1}{c|}{53.67}                                     & \cellcolor{pink}66.33  & \cellcolor{pink}70.33   & \multicolumn{1}{c|}{48.00}  & \cellcolor{pink}74.67     & 57.67     \\\cline{2-14}
                                 & Llama-1B                & \multicolumn{1}{c|}{48.00}                                     & 47.00  & 46.33   & \multicolumn{1}{c|}{\cellcolor{cyan}42.00}  & 53.67  & 49.33  & \multicolumn{1}{c|}{37.67}                                        & \cellcolor{cyan}28.33     & 36.33      & \multicolumn{1}{c|}{\cellcolor{cyan}17.33}     & \cellcolor{cyan}26.67     & \cellcolor{cyan}20.33     \\\hline
\end{NiceTabular}

\caption{Accuracy on SAQ (baseline), MCQ, and TFQ question types. The ``True'' and ``False'' columns indicate that the correct answers are true and false, respectively. See Section~\ref{tablenotations} for the notations and significance of color highlights.}
\label{overallresults}
\end{table*}

\subsection{Models and Experiments}

We chose OpenAI GPT models (gpt-4o and gpt-3.5-turbo) to represent large closed-source LLMs. We selected Llama3 8B (llama-8B-instruct), Llama3.2 1B (llama-1B-instruct), and Gemma 7B (gemma-7B-instruct) to represent small open-source LLMs for the repeatability of experiments. The default configurations used greedy decoding (i.e., temperature = 0). For each reasoning problem and each configuration, an API call was requested to that LLM once. The reasoning and final outputs were saved. We did not apply self-consistency since we needed to analyze reasoning accuracy. The experiments were conducted from late October to early December of 2024.

\subsection{Statistical Tests and Notations \label{tablenotations}}
In all performance comparisons for RQ1, RQ2, RQ4, and RQ5, we used the two-sided paired Wilcoxon signed-rank test \citep{Wilcoxon1992} to check for statistical significance in LLM performance differences between two different question types. The baseline can be different for different performance comparisons. The highlighted cells (both cyan and pink) in all the tables indicate statistically significant differences from the baseline (p-values $<$ 0.05). \colorbox{cyan}{Cyan} and \colorbox{pink}{pink} colors highlight that the baseline is better or worse than the compared method, respectively. Different positions of the correct answers of MCQs (U, U-SEO, and oSEO) are described in Section~\ref{mcq-design}. In Tables~\ref{number}-\ref{yon}, the difference (denoted as $\Delta$) is the LLM performance of the baseline less that of the compared question type. Q and S for TFQs denote the question and the statement formats, respectively, where “+st” indicates the instruction that asks to solve the problem first before answering the question. The term ``LLMs'' used in analyzing experimental results refers to the specific LLMs under study.

\section{RQ1: Do question types impact the final selection accuracy of LLM? \label{RQ1}}

Table~\ref{overallresults} (top) shows FS accuracy for commonly used 5-option MCQs and TFQs with the question format.
Our analysis description focuses on the cases with {\it statistically significant differences}. 

\subsection{Quantitative reasoning with GSM8K300}


{\bf SAQ} vs {\bf MCQ}: FS accuracy on SAQ is better than MCQ in 53\% of the cases (8/15), but worse than MCQ in 13.3\% of the cases (2/15). 
SAQ wins over MCQ (o-SEO) in 80\% of the cases (4/5), suggesting the difficulty in choosing the correct option with SEO as the correct answer.

{\bf SAQ} vs {\bf TFQ}: FS accuracy between the two question types is statistically significant in 6 of 10 cases (60\%). Four of these cases have TFQ (False) winning over SAQ. See Table~\ref{overallresults} (top) GSM8K300.  Both GPT models perform worse on SAQs. 

{\bf MCQ} vs {\bf TFQ}: FS accuracy on TFQ (False) is at least 7.88 higher than that of MCQ (U) in the four cases indicated by $*$ in Table~\ref{overallresults}.

\subsection{Deductive reasoning with bAbI300}

{\bf SAQ} vs {\bf MCQ}: SAQ wins over MCQ and loses in as many statically significant cases as those on GSM8K300. However, most models perform worse on this dataset than on GSM8K300, for instance, GPT-3.5 on MCQ (U-SEO).


{\bf SAQ} vs {\bf TFQ}: FS accuracy of all LLMs is influenced positively or negatively with TFQs with ``False'' as the correct answer.


{\bf MCQ} vs {\bf TFQ}: The symbol $^\dagger$ in Table~\ref{overallresults} indicates cases where TFQ (False) wins over MCQ (U) by at least 20 percent.

\section{RQ2: Do question types impact the reasoning accuracy of LLM? \label{RQ2}}

Table~\ref{overallresults} (bottom) shows reasoning accuracy for 5-option MCQs and TFQs with the question format. 

\subsection{Quantitative reasoning with GSM8K300}

{\bf SAQ} vs {\bf MCQ}:  80\% of the cases (12/15) do not show statistically significant reasoning accuracy differences.
 Reasoning accuracy difference positively co-occurs with FS accuracy difference in only one case, Llama-1B on MCQ (oSEO).

{\bf SAQ} vs {\bf TFQ}: 70\% of the cases (7/10) are not significant. The reasoning accuracy mostly does not correlate with FS accuracy. When ``False'' is the correct answer, LLMs need not be as accurate to choose the final correct answer. Gemma shows the opposite trend, with SAQ winning over TFQ (False). 


{\bf MCQ} vs {\bf TFQ}: Differences in reasoning accuracy are less pronounced compared to FS accuracy.

\begin{figure*}[pthb] {
\centering
\includegraphics[width=1\textwidth]{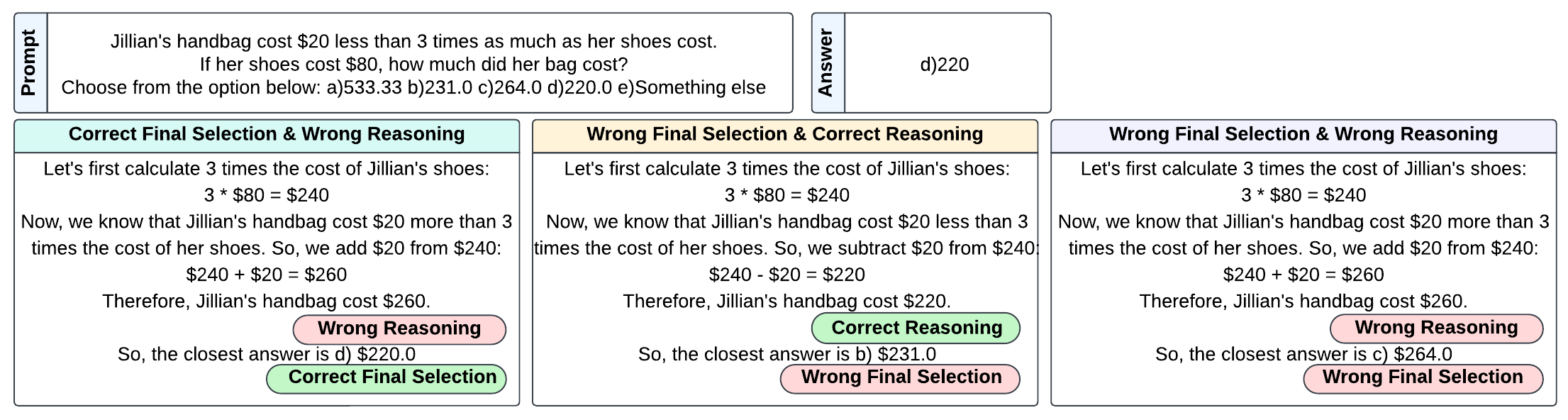} 
\caption{Examples of different patterns of incorrect outputs by LLMs for MCQ questions}
\label{incorrectoutput}}
\end{figure*}

\begin{figure}[h!tb] {
\centering
\includegraphics[width=0.5\textwidth]{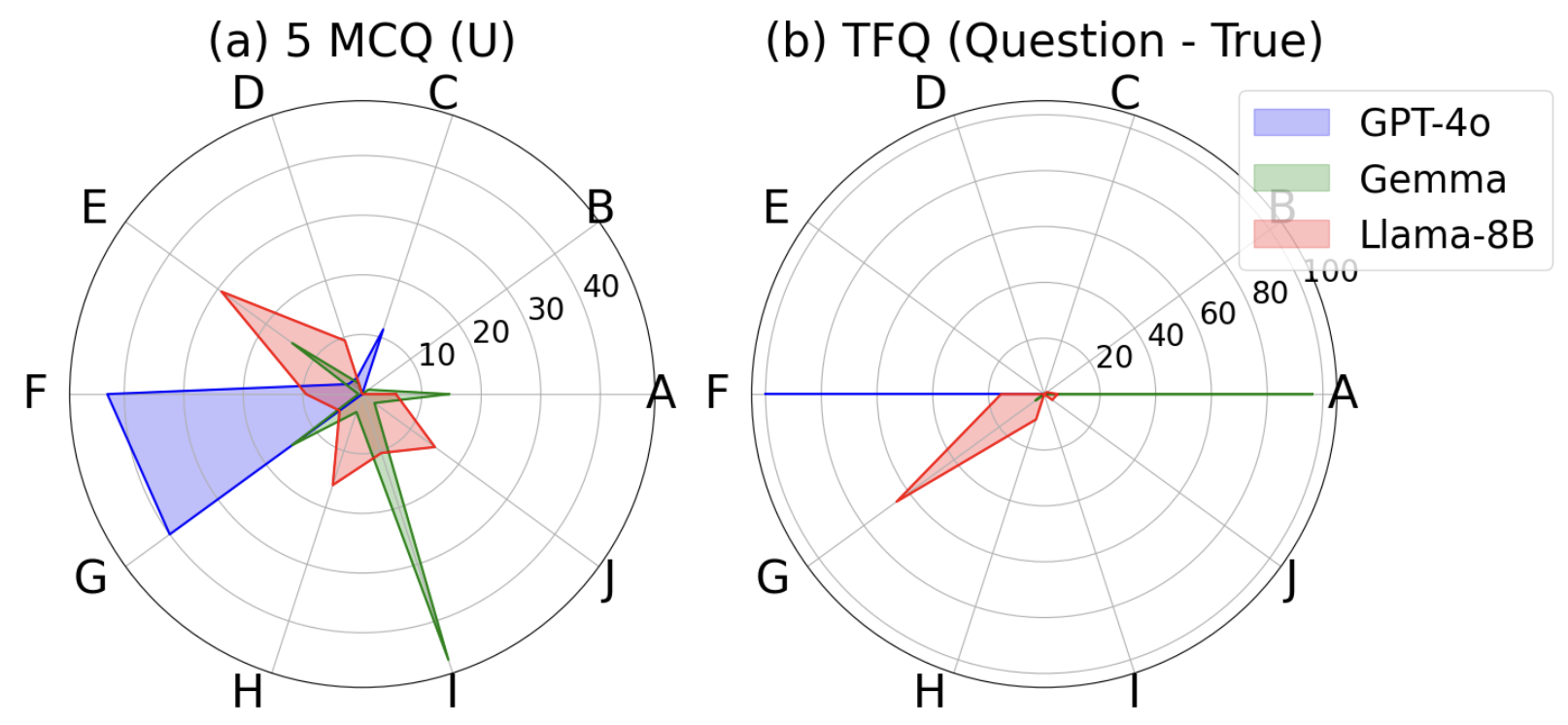} 
\caption{Percent of incorrect pattern outputs by LLMs (GPT-4o, Gemma, and Llama-8B) on (a) 5 MCQ (U) from 0 to 50\% and (b) TFQ (Question - True) from 0 to 100\%. A-J indicate the letters in front of each pattern.}
\label{radar}}
\end{figure}

\subsection{Deductive reasoning with bAbI300}

{\bf SAQ} vs {\bf MCQ}: SAQ wins over MCQ in 46.67\% (7/15) and loses in 20 \% (3/15) of the cases. Four cases of reasoning accuracy difference co-occur with FS accuracy difference.  

 {\bf SAQ} vs {\bf TFQ}: SAQ wins over MCQ in 50\% (5/10) and loses in 10\% (1/10) of the cases. Four cases of reasoning accuracy difference co-occur with FS accuracy difference.

{\bf MCQ} vs {\bf TFQ}: The notable case includes MCQ (U) winning over TFQ (False) by 32.67 on Gemma.

{\bf Summary:} Reasoning accuracy does not always correlate with FS accuracy.

\section{RQ3: What are the patterns of the incorrect outputs by LLMs on different question types?}

This analysis focuses only on incorrect output cases where either the final selection, the reasoning, or both are incorrect.
Figure~\ref{incorrectoutput} shows examples of different patterns. Our analysis reveals distinct patterns of mistakes for different types of questions. Figure~\ref{radar} shows statistics of the patterns by different models on 5-option MCQs (U) and TFQ (Question-False).

\subsection{Correct Final Selection but Wrong Reasoning}

The following patterns cause LLMs to perform better on MCQs or TFQs than SAQs.

\noindent \textbf{A. Good with guessing: } LLMs make incorrect reasoning steps but guess the final answer correctly for MCQs and TFQs. Regardless of whether ``True'' or ``False'' is the correct answer of TFQs, LLMs sometimes make a correct guess first before outputting their reasoning. This pattern occurs in approximately 96\% of incorrect outputs with Gemma on TFQ in Figure~\ref{radar}(b).

\noindent \textbf{B. Proximity helps correctness: } LLMs make nearly correct reasoning and calculations. With slightly off results, LLMs can still choose the correct option. This pattern appears in approximately 1\% of incorrect outputs, with Gemma on MCQs and Llama-8B on TFQs.

\noindent \textbf{C. Advantage of ``Something else'':} When this option is the correct answer, LLMs choose correctly sometimes, even with wrong reasoning steps. Around 11\% of incorrect outputs from GPT-4o on MCQs are in this category, as shown in Figure~\ref{radar}(a).

\subsection{Wrong Final Selection but Correct Reasoning}

LLMs sometimes return the wrong final selection after the correct reasoning, which leads to better LLM performance on SAQs than the other question types. We categorize the failure cases as follows.

\noindent \textbf{D. Incorrect option selection: } LLMs perform a valid calculation but choose the wrong option due to incorrect decimal or unit conversion for MCQs or an incorrect guess on TFQs before reasoning. Figure~\ref{radar}(a) shows approximately 9\% of Llama's outputs on MCQs select incorrect options.

\noindent \textbf{E. Misunderstanding ``Something else'': } LLMs fail to select ``None of the above'' or ``Something else'' in MCQs when appropriate, repeatedly recalculating to match its output to a specific numerical option. Figure~\ref{radar}(a) shows Gemma and Llama misunderstand ``SEO'' in approximately 29\% and 14\% of the outputs on MCQs, respectively.

\noindent \textbf{F. Missing final selection output: }LLMs sometimes fail to select the correct option despite completing the correct reasoning steps. This pattern occurs primarily with GPT-4o on MCQs and TFQs.

\subsection{Wrong Final Selection and Wrong Reasoning}

The failure cases are categorized as follows.

\noindent \textbf{G. Faulty reasoning leading to incorrect answers: }Due to errors in the reasoning or calculation steps, LLMs arrive at an incorrect final answer. Figure~\ref{radar}(a) shows approximately 40\% by GPT4o on MCQ and 65\% by Llama on TFQ outputs.

\noindent \textbf{H. Stopping in the middle of reasoning: }LLMs stop their reasoning process in the middle, resulting in incorrect reasoning steps and missing the final selection. Approximately 10\% of Llama's outputs have this pattern on both question types.

\noindent \textbf{I. Selecting different wrong options with wrong reasoning steps: }LLMs have wrong reasoning steps, leading to an incorrect answer. This pattern occurs primarily around 47\% with Gemma on MCQs.

{\noindent \textbf{J. Selecting nothing with wrong reasoning steps: } Only 5\% of LLama's outputs on MCQs are this pattern.}

\section{RQ4: What factors of MCQs influence LLM performance?}

\subsection{Positions of Correct Answers}

Table~\ref{overallresults} shows that the LLM performances fluctuate depending on the position of the correct answers. For most cases, except for GPT-4o and Llama-1B, the models perform best without SEO as a correct option (U-SEO).  Using only SEOs as the correct answers (oSEO) results in a significant performance drop on MCQs. The models sometimes selected SEOs as their final selection but struggled to select SEOs when their final reasoning answer did not match one of the provided options. They often chose the option with the number closest to their computed output rather than selecting SEOs.

\subsection{Number of Options}

 Using 2, 3, or 4 options in MCQs does not significantly influence accuracy \citep{order}. We investigate whether a larger number of options would change the finding on the FS accuracy. To assess the statistical significance of differences between the accuracies of 5 and 11-option MCQs, Recall the statistical test discussion in Section~\ref{tablenotations}. 
{\bf GSM8K300:} 
 Table \ref{number} shows no statistically significant differences between the 5-option and 11-option MCQs for GPT-4o and the Llama models. For the other models, the FS accuracy on the 5-option MCQs is better in 4 out of 12 cases. Gemma shows the largest absolute differences in both FS and reasoning accuracy. The 5-option MCQs clearly win on Gemma with and without ``Something else'' as the correct answer.
 {\bf bAbI300:} 
 All LLMs exhibit statistically significant differences in FS accuracy. The 5-option MCQs generally win over  11-option MCQs. Additionally, Llama-8B does not show a statistically significant difference in reasoning accuracy.

\subsection{Something else vs None of the above}

For this set of experiments, we replaced ``Something else'' with ``None of the above'' option to investigate the impact of word choices. Table \ref{none} shows the FS accuracy of GPT-3.5 improves in all cases with ``None of the above'' as the final option. However, the statistically significant improvements are limited to the 5-option MCQs when SEO is the correct answer for one-fifth of the reasoning problems (U) and all the problems (oSEO). These results indicate that LLMs are sensitive to variations in the wording of MCQ options.

\begin{table}[]
\centering
\small
\begin{NiceTabular}{llcccc}
                         &           & \multicolumn{2}{c}{GSM8K300} & \multicolumn{2}{c}{bAbI300}               \\ \cline{3-6} 
\multirow{-2}{*}{Models} & \multirow{-2}{*}{ } 
& $\Delta$ FS & $\Delta$ R  & $\Delta$ FS & $\Delta$ R         \\ \hline
& U        & -1.22 & 0.00  & \cellcolor{cyan}4.67  & -1.00        \\
& U-SEO       & -0.34 & 0.00  & 3.00  & \cellcolor{pink}-4.33         \\
\multirow{-3}{*}{GPT-4o}    
& oSEO      & 2.00  & -0.73  & \cellcolor{cyan}7.67  & 1.81       \\\hline

& U        & 3.34 & 0.00   & \cellcolor{pink}-7.67  & \cellcolor{cyan}6.00  \\
& U-SEO       & \cellcolor{cyan}12.00 & 1.67  & 1.66  & 2.33    \\
\multirow{-3}{*}{GPT-3.5}    
& oSEO      & \cellcolor{cyan}5.33  & 0.33  & \cellcolor{cyan}8.00  & \cellcolor{cyan}17.00    \\\hline

& U        & \cellcolor{cyan}12.00 & -4.82  & \cellcolor{cyan}16.66  & \cellcolor{cyan}16.00         \\
& U-SEO       & \cellcolor{cyan}21.66 & \cellcolor{cyan}5.00  & \cellcolor{cyan}26.00  &\cellcolor{cyan}19.66   \\
\multirow{-3}{*}{Gemma}    & oSEO      & -2.33  & -0.67  & -5.33  & \cellcolor{pink}-2.66      \\\hline
                         
& U       & 5.00 & 1.00  & 1.00  & 2.00   \\
 & U-SEO       & 3.00 & -1.33  & \cellcolor{cyan}6.00  & -1.33         \\
\multirow{-3}{*}{Llama-8B}  
& oSEO      & 5.00 & -0.33  & \cellcolor{cyan}6.00  & 4.33         \\\hline

& U      & 3.34  & 2.00   & \cellcolor{cyan}12.34  & 2.00         \\
& U-SEO       & 3.67  & 0.66  & 7.00  & 5.66         \\
\multirow{-3}{*}{Llama-1B}  
& oSEO      & 4.33 & -2.00  & \cellcolor{cyan}21.34  & \cellcolor{pink}-5.67  \\\hline
\end{NiceTabular}
\caption{Accuracy differences on 5-option MCQs (baseline) vs 11-option MCQs. Recall the notations and significance of color highlights in Section~\ref{tablenotations}.}
\label{number}
\end{table}

\begin{table}[]
\centering
\small
\begin{NiceTabular}{clccc}
\#Options       &   & \begin{tabular}[c]{@{}c@{}}Something \\ else\end{tabular} & \begin{tabular}[c]{@{}c@{}}None of \\ the above\end{tabular} & $\Delta$ FS                          \\ \hline
                         & U    & 69.67                                                     & 74.33                                                        & \cellcolor{pink}-4.66     \\
                         & U-* & 83.67                                                     & 85.33                                                        & -1.66                            \\
\multirow{-3}{*}{5 opts}  & o*    & 16.67                                                     & 32.33                                                        & \cellcolor{pink}-15.66 \\ \hline
                         & U    & 66.33                                                     & 68.00                                                        & -1.67                             \\
                         & U-* & 71.67                                                     & 73.67                                                        & -2.00                             \\
\multirow{-3}{*}{11 opts} & o*    & 11.33                                                     & 12.00                                                        & -0.67                            \\
\hline
\end{NiceTabular}
\caption{FS accuracy differences of GPT-3.5 on MCQs from GSM8K300 with ``Something else'' (baseline) vs ``None of the above'' as the final option; ``*'' indicates either SEO or ``None of the above.'' Recall the notations and significance of color highlights in Section~\ref{tablenotations}.}
\label{none}
\end{table}

\section{RQ5: What factors of TFQs influence LLMs' performance?}

\subsection{Question Format vs Statement Format}

Table \ref{instruction} shows influences of different TFQ formats. FS accuracy is worse for the question format than that of the statement format, except for GPT4-o on bAbI300 and Gemma on GSM8K300.
Reasoning accuracy differences do not always correlate with FS accuracy differences.

\subsection{True vs False as Correct Answers}

Table \ref{negation} shows the results.
 {\bf GSM8K300: }On FS accuracy, GPT-3.5 and Llama-8B tend to perform significantly better with ``False'' as correct answers, while Gemma and Llama-1B perform better on TFQs with ``True'' as correct answers. For GPT4-o, there are no significant differences in reasoning and FS accuracy.
{\bf bAbI300: } On FS accuracy, GPT models tend to perform significantly better with ``False'' as correct answers, while Gemma performs better on TFQs with ``True'' as correct answers. LLMs mostly reason better with TFQs with ``True'' as correct answers. 

\begin{table}[]
\centering
\small
\begin{tabular}{llcccccc} 
                         &           & \multicolumn{2}{c}{GSM8K300} & \multicolumn{2}{c}{bAbI300}               \\ \cline{3-6} 
\multirow{-2}{*}{Models} & \multirow{-2}{*}{ } 
& $\Delta$ FS & $\Delta$ R  & $\Delta$ FS & $\Delta$ R         \\ \hline
& Q-S      & 0.67 & 1.33  & \cellcolor{cyan}11.67  & -2.00       \\
\multirow{-2}{*}{GPT-4o}    
& Q-S+st      & -0.67  & -0.98  & -9.66  & \cellcolor{pink}-10.00      \\\hline

& Q-S       & -1.33 & -2.67  & \cellcolor{pink}-50.67  & \cellcolor{pink}-29.00  \\
\multirow{-2}{*}{GPT-3.5}    
& Q-S+st      & -2.00  & -1.34  & \cellcolor{pink}-33.00  & \cellcolor{pink}-24.34   \\\hline

& Q-S       & \cellcolor{cyan}5.33 & -1.67  & -0.67  & \cellcolor{pink}-52.33  \\
\multirow{-2}{*}{Gemma}    
& Q-S+st      & \cellcolor{pink}-12.33  & \cellcolor{pink}-6.13  & \cellcolor{pink}-32.00  & \cellcolor{pink}-55.66     \\\hline
                         
 & Q-S       & \cellcolor{pink}-4.66 & -3.33  & -1.00  & \cellcolor{pink}-7.00  \\
\multirow{-2}{*}{Llama-8B}  
& Q-S+st      & \cellcolor{pink}-15.00 & -3.00  & \cellcolor{pink}-22.00  & \cellcolor{pink}-21.33      \\\hline

& Q-S       & \cellcolor{pink}-7.66  & 1.67  & \cellcolor{pink}-11.67  & \cellcolor{pink}-30.00       \\
\multirow{-2}{*}{Llama-1B}  
& Q-S +st     & \cellcolor{pink}-4.33 & -0.66  & \cellcolor{pink}-6.67  & \cellcolor{pink}-10.33  \\
\hline
\end{tabular}
\caption{Accuracy differences on TFQs using ``True'' as the correct answer of the question format (Q) (baseline) vs the statement format (S). Recall the notations and significance of color highlights in Section~\ref{tablenotations}.}
\label{instruction}
\end{table}

\subsection{True or False vs Yes or No}

Are LLMs sensitive to the choice of words for TFQs? In this set of experiments, we replaced ``True or False'' with ``Yes or No'' in the prompts, keeping everything else identical to the original TFQs prompts. Table \ref{yon} shows the FS accuracies of GPT-3.5 drops with statistical significance when ``Yes or No'' instead of ``True or False'' in all scenarios except when the question format is used together with the guiding instruction to solve the problem first (+st). Using ``Yes or No'' has a performance drop of 15 and 26.67\% without the guiding instruction but is only around 0.33\% otherwise. 


\begin{table}[]
\centering
\small
\begin{NiceTabular}{llcccccc} 
                         &           & \multicolumn{2}{c}{GSM8K300} & \multicolumn{2}{c}{bAbI300}               \\ \cline{3-6} 
\multirow{-2}{*}{Models} & \multirow{-2}{*}{ } & $\Delta$ FS & $\Delta$ R  & $\Delta$ FS & $\Delta$ R         \\ \hline

& Q        & -4.33 & 1.00  & \cellcolor{pink}-29.33  & \cellcolor{pink}-9.00  \\
                         & Q+st       & -4.00 & 0.34  & \cellcolor{pink}-15.33  & \cellcolor{cyan}17.33  \\
\multirow{-2}{*}{GPT-4o}    & S      & -5.00  & 0.02  & \cellcolor{pink}-43.67 & \cellcolor{pink}-14.67 \\
& S+st      & -3.33  & 0.65  & 2.33  & \cellcolor{cyan}27.67 \\\hline

& Q        & \cellcolor{pink}-21.00 & -1.33  & \cellcolor{pink}-76.34  & -5.00  \\
                         & Q+st       & \cellcolor{pink}-16.00 & 0.00  & \cellcolor{pink}-27.67  & \cellcolor{cyan}16.00  \\
\multirow{-2}{*}{GPT-3.5}    & S      & \cellcolor{pink}-18.33  & 1.34  & \cellcolor{pink}-20.33  & \cellcolor{cyan}26.34  \\
& S+st      & \cellcolor{pink}-13.67  & 3.34  & 4.00  & \cellcolor{cyan}67.34  \\\hline

& Q        & \cellcolor{cyan}88.33 & \cellcolor{cyan}11.33  & \cellcolor{cyan}80.33  & \cellcolor{cyan}28.00  \\
                         & Q+st      & \cellcolor{cyan}12.67 & 3.87  & \cellcolor{cyan}33.67  & \cellcolor{cyan}18.00  \\
\multirow{-2}{*}{Gemma}    & S      & \cellcolor{cyan}69.67  & \cellcolor{cyan}23.67  & \cellcolor{cyan}82.67  & \cellcolor{cyan}79.00  \\
& S+st      & \cellcolor{cyan}60.33  & \cellcolor{cyan}67.00  & \cellcolor{cyan}99.00  & \cellcolor{cyan}89.33  \\\hline

& Q        & \cellcolor{pink}-12.66 & -0.66  & \cellcolor{pink}-12.67  & \cellcolor{cyan}17.00  \\
                         & Q+st       & \cellcolor{pink}-18.00 & 1.67  & \cellcolor{pink}-18.67  & \cellcolor{cyan}15.67  \\
\multirow{-2}{*}{Llama-8B}    & S      & -2.34  & 3.67  & -6.00  & \cellcolor{cyan}36.00  \\
& S+st      & \cellcolor{cyan}9.33  & \cellcolor{cyan}5.34  & \cellcolor{cyan}15.33  & \cellcolor{cyan}44.66  \\\hline

& Q        & \cellcolor{cyan}10.67 & 4.34  & \cellcolor{pink}-8.67  & 6.34  \\
                         & Q+st       & \cellcolor{cyan}34.67 & 2.34  & 1.33  & \cellcolor{cyan}31.67  \\
\multirow{-2}{*}{Llama-1B}    & S      & \cellcolor{cyan}8.33  & \cellcolor{cyan}15.00  & 0.33  & \cellcolor{cyan}30.34  \\
& S+st     & \cellcolor{cyan}44.00  & \cellcolor{cyan}12.00  & 3.67  & \cellcolor{cyan}30.00  \\
\hline
\end{NiceTabular}
\caption{Accuracy differences on TFQs with ``True'' (baseline) vs ``False'' as the correct answers. Recall the notations and significance of color highlights in Section~\ref{tablenotations}.}
\label{negation}
\end{table}

\begin{table}[!htbp]
\centering
\small
\begin{NiceTabular}{cccc}
Types  & True or False    & Yes or No    & $\Delta$ FS                           \\ \hline
Q      & 75.67 & 49.00 & \cellcolor{cyan}26.67 \\
Q+st & 80.33 & 80.00 & 0.33                             \\
S      & 77.00 & 62.00 & \cellcolor{cyan}15.00 \\
S+st & 82.33 & 87.67 & \cellcolor{pink}-5.34    \\
\hline
\end{NiceTabular}
\caption{GPT-3.5's FS accuracy on TFQs between ``True or False'' (baseline) and ``Yes or No'' using GSM8K300. Recall the notations and significance of color highlights in Section~\ref{tablenotations}.}
\label{yon}
\end{table}

\section{Conclusions and Future Work}

LLM performance on different question types fluctuates. Our experimental results show that the reasoning accuracy is not always correlated with the final selection accuracy for multiple-choice questions and True or False questions. To improve LLM performance on these question types, it is important to improve both reasoning accuracy and the accuracy of selecting the final answer. The trends in significant differences vary across different reasoning tasks. For MCQs, factors such as the position of correct answers, the number of options, and the selection of words in options impact LLM performance. In TFQs, factors such as type of instruction, whether ``True'' or ``False'' is the correct answer, and the selection of words influence LLM performance.

\section{Limitations}

This paper presents a comprehensive evaluation of the impact of different question types on LLM performance, but with the following limitations. The experiments do not involve prompting methods or few-shot learning strategies. The proposed measure of reasoning accuracy does not account for accuracy at individual reasoning steps, nor does it consider how the number of reasoning steps affects performance.  While deeper insights could be obtained through such a study, it would require substantially more manual labeling effort. Lastly, due to computational constraints, the experiments were conducted on subsets of the two datasets. Nevertheless, we conducted statistical tests to evaluate the impact of question types on LLM performance.

\section{Ethical Considerations and Potential Risk}

The findings presented in this study are based on experiments conducted on five different large language models for quantitative and deductive reasoning tasks on standard open-source datasets under a zero-shot setting. Each of these language models has been trained on a distinct corpus with a specific training objective, making the obtained outputs dependent on the experimental setup. Our objective is to empirically demonstrate that LLM performance fluctuates across different question types. This variation highlights the need to improve both reasoning accuracy and the accuracy of selecting the final answer to enhance overall performance. Consequently, the results and analysis may not generalize to other large language models or to the same models if fine-tuned on the same or different datasets.

\section{Acknowledgement}

This work is partially supported by the NSF Grant No. 2152117. The findings, interpretations, and conclusions presented in this paper are those of the authors and do not necessarily reflect the views of the funding agency. The authors gratefully acknowledge Owen Harty, Vicky Lee, Noah Shpardson, and Yonas Sium for their invaluable assistance with annotation.

\bibliography{custom}

\appendix

\section{Appendix}
\label{sec:appendix}

\subsection{Related Work about Datasets}
LLMs have been tested with various reasoning tasks: mathematical reasoning, deductive reasoning, causal reasoning, visual reasoning, and commonsense reasoning, just to name a few~\citep{qiao-etal-2023-reasoning}. Benchmarking is one of the most widely adopted methods to evaluate LLM performance on these tasks. We list existing datasets on two reasoning tasks and question types used in this study.

\subsubsection{Quantitative and Deductive Reasoning Tasks}
Math Word Problems datasets (e.g., MAWPS \citep{mawps}, SVAMP \citep{svamp}, and GSM8K \citep{gsm}) are commonly used benchmarks for quantitative reasoning tasks. Recent datasets, e.g., GSM-IC \citep{gsm-ic} and MPN \citep{scott}), include irrelevant information to increase reasoning complexity.
Some benchmarks target classical logical reasoning, including deductive reasoning tasks. Example datasets include reasoning benchmark in philosophy and logic \citep{reasoning_survey}, bAbI \citep{weston2016babi}, RuleTaker \citep{ruletaker}, ProofWriter \citep{proofwriter}, and RobustLR \citep{robustlr}.

\subsubsection{Question Types} Question types can be broadly categorized into short/long answer questions, multiple-choice questions, and true/false or yes/no questions. We reviewed existing LLM benchmark datasets listed in the Github\footnote{https://github.com/leobeeson/llm\_benchmarks}. Most of the datasets have only one question type. A few datasets \citep{Bisk2019PIQARA, nguyen2016ms} have more than one question type.
Multiple-choice questions appear most frequently. True/false or yes/no questions appear least often. Short/long answer questions are included in diverse benchmarks for Math Word Problems, reading comprehension, and language understanding. 

Examples of datasets that include different question types are listed below. 

\noindent {\bf Multiple-choice questions:} MMLU~\citep{hendryckstest2021}, ARC~\citep{ARC2021}, HellaSwag~\citep{zellers-etal-2019-hellaswag}, MultiNLI~\citep{williams-etal-2018-broad}, WinoGrande~\citep{sakaguchi-winnogrande-2021}, PDP~\citep{Morgenstern_Davis_Ortiz_pdp-2016}, SuperGLUE-WSC~\citep{sakaguchi-winnogrande-2021}, DPR~\citep{rahman-ng-2012-resolving}, KnowRef~\citep{emami-etal-2019-knowref}, COPA~\citep{Gordon2011ChoiceOP}, Winogender~\citep{rudinger-EtAl:2018:N18}, SciQ~\citep{welbl-etal-2017-crowdsourcing}, CRASS~\citep{frohberg-binder-2022-crass}, PIQA~\citep{Bisk2019PIQARA}, CommonsenseQA, Social IQa, HellaSWAG~\citep{zellers-etal-2019-hellaswag}, and Numersense \citep{numersenselin-etal-2020-birds}. Three to five options are common, whereas two and eleven options are less common.

\noindent {\bf Short/long answer questions:} GSM8K~\citep{gsm}, SVAMP, MAWPS, Natural Questions~\citep{kwiatkowski-etal-2019-natural}, GLUE~\citep{wang-etal-2018-glue}, LAMBADA~\citep{paperno-etal-2016-lambada}, SuperGLUE~\citep{wang-superglue-2019}, TriviaQA~\citep{joshi-etal-2017-triviaqa}, DROP~\citep{dua-etal-2019-drop}, MS-MARCO~\citep{nguyen2016ms}, and PIQA.

\noindent {\bf True/false or yes/no  questions:} BoolQ~\citep{clark-etal-2019-boolq}, MS-MARCO~\citep{nguyen2016ms}.

Most existing datasets mostly adopt one question type. A few have more than one question type: PIQA and MS-MARCO.

\subsection{Methods to Generate Wrong Options}

To generate wrong answers for the deductive reasoning task, our program randomly picked three values from the five possible wrong answers for the 5-option MCQs. These values are `kitchen', `bedroom', `garden', `bathroom', `hallway', and `office'. For the 11 option MCQs, nine values were randomly selected from the eleven values of wrong answers, including `beds', `baths', `table', `desk', and `livingroom' in addition to the five words used to generate wrong answers for the 5-option MCQs.

When SEO is the correct answer, the ground truth needs to be removed from the options. For the quantitative reasoning task, we randomly chose a number with the same number of digits. While four or ten values are selected from the possible answers for the deductive reasoning task.

\subsection{Examples of Datasets}

See Tables 7 and 8.

\begin{table*}[!htbp]
\centering
\begin{tabular}{l}
\hline
\textbf{Question: }\\Chatty prepared 3 dozen eggs for her 4 children's Easter activity. \\ Assuming each child gets the same number of eggs, how many eggs does each child receive? \\
\textbf{Ground Truth: }23\\
\hline
\end{tabular}
\caption{Example of GSM8K300}
\end{table*}

\begin{table*}[!htbp]
\centering
\begin{tabular}{l}
\hline
\textbf{Question: }\\Daniel journeyed to the office. John traveled to the office. \\ Mary went back to the garden. Mary went to the bedroom. Mary put down the milk. \\Where was the milk before the bedroom?\\
\textbf{Ground Truth: }garden\\
\hline
\end{tabular}
\caption{Example of bAbI300}
\end{table*}

\subsection{Prompt Designs to Format Question Types}

See Tables 9 to 14.

\begin{table*}[!htbp]
\centering
\begin{tabular}{l}
\hline
\textbf{Prompt: }\\Chatty prepared 3 dozen eggs for her 4 children's Easter activity. \\ Assuming each child gets the same number of eggs, how many eggs does each child receive? \\
\textbf{Output from GPT4o: }\\Chatty prepared 3 dozen eggs. \\Since 1 dozen equals 12 eggs, 3 dozen eggs is:\\

$3 \times 12 = 36 \text{ eggs}$ \\

She has 4 children, and each child should receive the same number of eggs. \\To find out how many eggs each child receives,\\ divide the total number of eggs by the number of children:\\

$\frac{36}{4} = 9 $\\

Each child receives 9 eggs.\\
\hline
\end{tabular}
\caption{Example of SQA}
\end{table*}

\begin{table*}[!htbp]
\centering
\begin{tabular}{l}
\hline
\textbf{Prompt:} \\Chatty prepared 3 dozen eggs for her 4 children's Easter activity. \\ Assuming each child gets the same number of eggs, how many eggs does each child receive? \\Chose from the option below: \\
a)41.0 b)27.6 c)24.15 d)1350.0 e)Something else \\
\textbf{Output from GPT4o: }\\
Let's solve the problem step by step.\\

1. **Red Cars**: The rink has 12 red cars.\\

2. **Green Cars**: \\They have 2 fewer green cars than red cars. So, the number of green cars is:\\
   $
   12 - 2 = 10
   $\\

3. **Blue Cars**: \\They have 3 times the number of blue cars as green cars. So, the number of blue cars is:\\
   $
   3 \times 10 = 30
   $\\

4. **Total Cars**: \\The total number of cars is 75.\\

5. **Yellow Cars**: \\Let $ y $ be the number of yellow cars. The equation for the total number of cars is:\\
   $
   12 + 10 + 30 + y = 75
   $\\

   Simplifying this equation:\\
   $
   52 + y = 75
   $\\

   Solving for $y$:\\
   
   $y = 75 - 52 = 23$\\

So, the rink has 23 yellow cars. The correct answer is "e) Something else".\\

\hline
\end{tabular}
\caption{Example of MCQ}
\end{table*}

\begin{table*}[!htbp]
\centering
\begin{tabular}{l}
\hline
\textbf{Prompt <Question>:} \\
Chatty prepared 3 dozen eggs for her 4 children's Easter activity. \\ Assuming each child gets the same number of eggs, how many eggs does each child receive? \\
Is the answer 23? Choose one of these. True or False\\
\textbf{Output from GPT4o: }\\
True. \\

Chatty prepared 3 dozen eggs, which is equal to $3 \times 12 = 36$ eggs. \\If these 36 eggs are distributed equally among her 4 children, each child would receive:\\

$\frac{36}{4} = 9$\\

So, each child receives 9 eggs. Therefore, the statement "each child receives 9 eggs" is true.\\
\hline
\end{tabular}
\caption{Example of TFQ Question}
\end{table*}

\begin{table*}[!htbp]
\centering
\begin{tabular}{l}
\hline
\textbf{Prompt <Question + Step instruction>:} \\
Chatty prepared 3 dozen eggs for her 4 children's Easter activity. \\ Assuming each child gets the same number of eggs, how many eggs does each child receive? \\
Is the answer 23? Solve the question first and choose. True or False\\
\textbf{Output from GPT4o: }\\
To solve the problem, we first need to determine the total number of eggs. \\Chatty prepared 3 dozen eggs. Since one dozen equals 12 eggs, 3 dozen eggs is:\\

 $3 \times 12 = 36 \text{ eggs}$\\

Chatty has 4 children, and each child should receive the same number of eggs. \\To find out how many eggs each child receives, \\we divide the total number of eggs by the number of children:\\

$\frac{36}{4} = 9$\\

So, each child receives 9 eggs. \\Therefore, the statement "Is the answer 9?" is True.\\

\hline
\end{tabular}
\caption{Example of TFQ Question + Step instruction}
\end{table*}

\begin{table*}[!htbp]
\centering
\begin{tabular}{l}
\hline
\textbf{Prompt <Statement>:} \\
Chatty prepared 3 dozen eggs for her 4 children's Easter activity. \\ Assuming each child gets the same number of eggs, how many eggs does each child receive? \\
The answer of the question is 23. True or False\\
\textbf{Output from GPT4o: }\\
True. \\

Chatty prepared 3 dozen eggs, which is equal to $3 \times 12 = 36$ eggs.\\ If these 36 eggs are divided equally among her 4 children, each child would receive:\\

$\frac{36}{4} = 9$\\

Therefore, each child receives 9 eggs. \\The statement that each child receives 9 eggs is true.\\

\hline
\end{tabular}
\caption{Example of TFQ Statement}
\end{table*}

\begin{table*}[!htbp]
\centering
\begin{tabular}{l}
\hline
\textbf{Prompt <Statement + Step instruction>:} \\
Chatty prepared 3 dozen eggs for her 4 children's Easter activity. \\ Assuming each child gets the same number of eggs, how many eggs does each child receive? \\
The answer of the question is 23. Solve the question first and choose. True or False\\
\textbf{Output from GPT4o: }\\
To solve the problem, we start by determining the total number of eggs Chatty prepared.\\ Since 1 dozen equals 12 eggs, 3 dozen eggs is:\\

$3 \times 12 = 36 \text{ eggs}$\\

Chatty has 4 children, and each child should receive the same number of eggs. \\To find out how many eggs each child receives,\\ we divide the total number of eggs by the number of children:\\

$\frac{36}{4} = 9 \text{ eggs per child}$\\

Therefore, each child receives 9 eggs. \\The statement that the answer to the question is 9 is true. \\

So, the answer is: True.\\
\hline
\end{tabular}
\caption{Example of TFQ Statement + Step instruction}
\end{table*}

\begin{table*}[!htbp]
\centering
\scriptsize
\renewcommand{\arraystretch}{1.5} 
\setlength{\tabcolsep}{3pt} 
\begin{NiceTabular}{llccccccccccccccc}
\multirow{3}{*}{\rotatebox{90}{Accuracy}} & \multirow{3}{*}{Models} & \multicolumn{1}{|c|}{\multirow{3}{*}{SAQ}} & \multicolumn{6}{c|}{MCQ}  & \multicolumn{8}{c}{TFQ}   \\ \cline{4-17}  
&                         & \multicolumn{1}{|c|}{}                     & \multicolumn{3}{c|}{5 options}           & \multicolumn{3}{c|}{11 options}          & \multicolumn{4}{c|}{True as correct answer}   & \multicolumn{4}{c}{False as correct answer} \\ \cline{4-17}   
&                         & \multicolumn{1}{|c|}{(baseline)}                     & U & U-SEO & \multicolumn{1}{c|}{oSEO} & U & U-SEO & \multicolumn{1}{c|}{oSEO} & Q  & Q+st & S  & \multicolumn{1}{c|}{S+st} & Q  & Q+st & S  & S+st \\ \hline
\multirow{5}{*}{\rotatebox{90}{Final Selection}}

& GPT-4o                 & \multicolumn{1}{|c|}{92.00}                                   & 89.45  & 91.33   & 97.67                         & 91.67  & 91.67  & 95.67                       & 93.00  & 93.33     & 92.33 & 94.00                         & 97.33 & 97.33     & 97.33 & 97.33     \\\cline{2-17}

& GPT-3.5                 & \multicolumn{1}{|c|}{79.60}                                         & 69.67  & 83.67   & 16.67                         & 66.33  & 71.67   & 11.33                         & 75.67  & 80.33      & 77.00 & 82.33                          & 96.67 & 96.33     & 95.33 & 96.00     \\\cline{2-17} 
                         & Gemma                  & \multicolumn{1}{|c|}{58.67} & 54.33                                        & 67.33 & 2.00  & 42.33                        & 45.67 & 4.33                        & 98.33 & 77.67     & 93.00 & 90.00                          & 10.00 & 65.00     & 23.33 & 29.67     \\\cline{2-17} 
                         & Llama-8B                  & \multicolumn{1}{|c|}{75.33} & 66.67                                        & 77.33 & 19.00  & 61.67                        & 74.33 & 14.00                        & 73.67 & 64.33     & 78.33 & 79.33 & 86.33     & 82.33 & 80.67 & 70.00     \\\cline{2-17} 

                         & Llama-1B                  & \multicolumn{1}{|c|}{48.00}    & 11.67 & 10.67  & 5.33   & 8.33 & 7.00  & 1.00                       & 47.67 & 65.00     & 55.33 & 69.33                          & 37.00* & 30.33     & 47.00 & 25.33
                         \\\hline
                         \hline

                       \multirow{5}{*}{\rotatebox{90}{Reasoning}}   

& GPT-4o                 & \multicolumn{1}{|c|}{92.00}                                         & 93.67  & 93.67   & 92.67     & 93.67  & 93.67  & 93.40                       & 95.00  & 93.67      & 93.67 & 94.65                    & 94.00 & 93.33     & 93.65 & 94.00    \\ \cline{2-17}

& GPT-3.5                 & \multicolumn{1}{|c|}{79.60}                                        & 81.00  & 80.67   & 80.00                         & 81.00  & 79.00  & 79.67                       & 78.00  & 81.33      & 80.67 & 82.67                    & 79.33 & 81.33     & 79.33 & 79.33    \\ \cline{2-17}  
                         & Gemma                  & \multicolumn{1}{|c|}{58.67} & 54.85                                        & 60.67 & 54.00  & 59.67                        & 55.67 & 54.67                        & 57.00 & 60.87     & 58.67 & 67.00                          & 45.67 & 57.00     & 35.00 & 46.67     \\ \cline{2-17} 
                         
                         & Llama-8B     & \multicolumn{1}{|c|}{75.33}    & 81.00  & 81.00  & 76.67   & 80.00 & 82.33  & 77.00   & 76.67 & 79.67     & 80.00 & 82.67 & 77.33 & 78.00 & 76.33 & 77.33   \\ \cline{2-17}

                         & Llama-1B                  & \multicolumn{1}{|c|}{48.00}   & 47.00  & 46.33   & 42.00                         & 45.00  & 45.67  & 44.00                       & 53.67  & 50.67      & 52.00 & 51.33                    & 49.33 & 46.00     & 37.00 & 39.33
                         \\ 
                         \hline

\end{NiceTabular}
\caption{Accuracy on SAQ (baseline), MCQ, and TFQ question types with GSM8K300. Recall the notations and significance of color highlights in Section~\ref{tablenotations}. }
\end{table*}

\begin{table*}[!htbp]
\label{bAbIresults}
\centering
\scriptsize
\renewcommand{\arraystretch}{1.5} 
\setlength{\tabcolsep}{3pt} 
\begin{NiceTabular}{lcccccccccccccccc}
\multirow{3}{*}{\rotatebox{90}{Accuracy}} & \multirow{3}{*}{Models} & \multicolumn{1}{c|}{\multirow{3}{*}{SAQ}} & \multicolumn{6}{c|}{MCQ}  & \multicolumn{8}{c}{TFQ}   \\ \cline{4-17}  
&                         & \multicolumn{1}{c|}{}                     & \multicolumn{3}{c|}{5 option}           & \multicolumn{3}{c|}{11 option}          & \multicolumn{4}{c|}{True}   & \multicolumn{4}{c}{False} \\ \cline{4-17}   
&                         & \multicolumn{1}{c|}{(baseline)}                     & U & U-SEO & \multicolumn{1}{c|}{oSEO} & U & U-SEO & \multicolumn{1}{c|}{oSEO} & Q  & Q+st & S  & \multicolumn{1}{c|}{S+st} & Q  & Q+st & S  & S+st \\ \hline
\multirow{5}{*}{\rotatebox{90}{Final Selection}}

& GPT-4o     & 82.33    & 79.00  & 80.00  & 83.67   & 64.33 & 77.00  & 76.00   & 67.67 & 82.67     & 56.00 & 92.33 & 97.00 & 98.00  & 99.67 & 90.00    \\ \cline{2-17}

& GPT-3.5                & 52.67                                         & 30.00  & 38.33   & 10.67                       & 37.67  & 36.67   & 2.67                         & 16.33  & 67.00      & 67.00 & 100.00                         & 92.67  & 94.67      & 87.33 & 96.00     \\ \cline{2-17} 
                         & Gemma                   & 36.33    & 39.33 & 49.67  & 0.00      & 22.67 & 23.67  & 5.33       & 99.33 & 68.00     & 100.00 & 100.00                          & 19.00 & 34.33     & 17.33 & 1.00    \\ \cline{2-17} 
                         & Llama-8B     & 53.67    & 59.00 & 69.67  & 21.00   & 58.00 & 63.67  & 15.00                        & 77.33 & 64.33     & 78.33 & 86.33                          & 88.00 & 83.00     & 84.33 & 71.00    \\ \cline{2-17}
                         & Llama-1B    & 37.67    & 20.67 & 14.33  & 23.67   & 8.33 & 7.33  & 2.33                        & 36.33 & 49.00     & 48.00 & 55.67                          & 45.00 & 47.67     & 39.67 & 52.00    \\
                         \hline 
                         \hline

\multirow{5}{*}{\rotatebox{90}{Reasoning}}   

& GPT-4o     & 82.33    & 74.33  & 81.67  & 41.81   & 75.33 & 86.00  & 40.00   & 59.33 & 82.67     & 61.33 & 92.67 & 68.33 & 65.33 & 76.00 & 65.00   \\ \cline{2-17}

& GPT-3.5                & 52.67                                         & 53.33  & 60.67  & 37.67   & 47.33 & 58.33  & 20.67   & 48.67 & 69.33     & 77.67 & 93.67 & 53.67 & 53.33 & 51.33 & 26.33   \\ \cline{2-17}
                         & Gemma                   & 36.33    & 38.00  & 45.30  & 0.67  & 22.00 & 25.67  & 3.33   & 30.67 & 34.67     & 83.00 & 90.33 & 2.67 & 16.67 & 4.00 & 1.00   \\ \cline{2-17}
                         & Llama-8B     & 53.67    & 66.33  & 70.33  & 48.00   & 64.33 & 67.33  & 43.67   & 74.67 & 66.00     & 81.67 & 87.33 & 57.67 & 50.33 & 45.67 & 42.67   \\ \cline{2-17}
                         & Llama-1B   &37.67 & 28.33  & 36.33  & 17.33   & 31.00 & 30.67  & 23.00   & 26.67 & 40.67     & 56.67 & 55.00 & 20.33 & 19.06 & 26.33 & 26.67   \\ \cline{2-17} \\
                         
\end{NiceTabular}
\caption{Accuracy on SAQ (baseline), MCQ, and TFQ question types with bAbI300. Recall the notations and significance of color highlights in Section~\ref{tablenotations}.}
\end{table*}

\subsection{Complete Accuracy Report for All Configurations Except for ``None of the above'' for MCQs and ``Yes or No'' for TFQs}

See Tables 15 to 16.

\end{document}